\documentclass[runningheads]{llncs}
\usepackage[T1]{fontenc}
\usepackage{graphicx}
\usepackage{algorithm}
\usepackage{algorithmic}
\usepackage{amsmath}
\usepackage{color}
\usepackage{multirow}
\usepackage{makecell}
\usepackage{hyperref}
\usepackage[misc]{ifsym}

\begin{document}
	\title{DARC: Distribution-Aware Re-Coloring Model for Generalizable Nucleus Segmentation}
	\author{
		Shengcong Chen\and
		Changxing Ding\and
		Dacheng Tao\and
		Hao Chen
	}
	\authorrunning{S. Chen et al.}
	
	\institute{ }

	\maketitle
	
	\begin{abstract}
		Nucleus segmentation is usually the first step in pathological image analysis tasks. Generalizable nucleus segmentation refers to the problem of training a segmentation model that is robust to domain gaps between the source and target domains. The domain gaps are usually believed to be caused by the varied image acquisition conditions, e.g., different scanners, tissues, or staining protocols. In this paper, we argue that domain gaps can also be caused by different foreground (nucleus)-background ratios, as this ratio significantly affects feature statistics that are critical to normalization layers. We propose a Distribution-Aware Re-Coloring (DARC) model that handles the above challenges from two perspectives. First, we introduce a re-coloring method that relieves dramatic image color variations between different domains. Second, we propose a new instance normalization method that is robust to the variation in foreground-background ratios. We evaluate the proposed methods on two H$\&$E stained image datasets, named CoNSeP and CPM17, and two IHC stained image datasets, called DeepLIIF and BC-DeepLIIF. Extensive experimental results justify the effectiveness of our proposed DARC model. Codes are available at \url{https://github.com/csccsccsccsc/DARC}.
	\end{abstract}
	
	\section{Introduction}
	
\begin{figure}[t]
		\centering
		\includegraphics[width=0.9\textwidth]{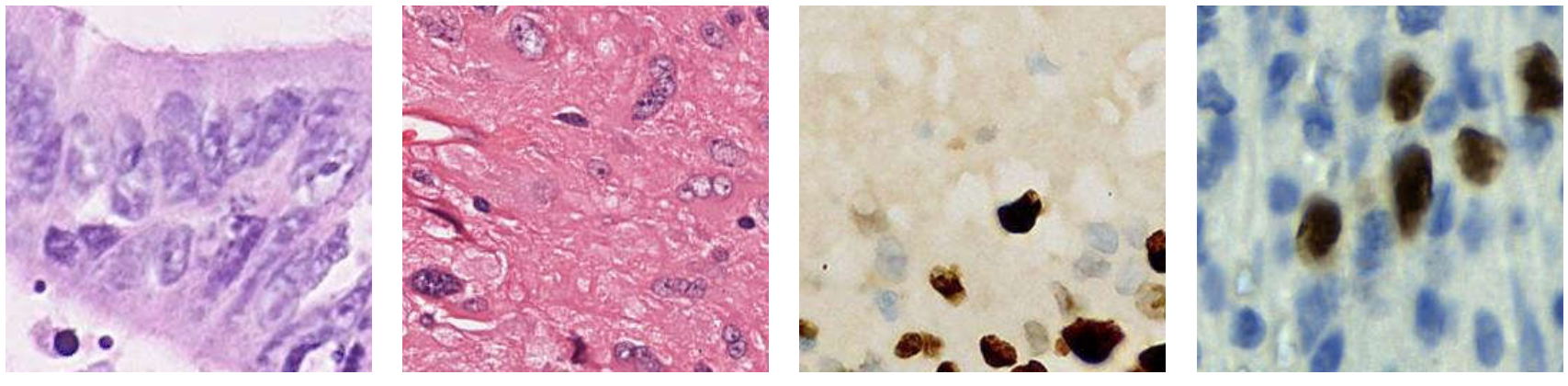}
		\label{fig:examples}
		\caption{Example image patches from different datasets. Their appearance differs significantly from each other due to variations in image modalities, staining protocols, scanner types, and tissues.}
	\end{figure}

	Automatic nucleus segmentation has captured wide research interests in recent years due to its importance in pathological image analysis \cite{dcan,cianet,hovernet,stardist}. However, as shown in Fig.~\ref{fig:examples}, the variations in image modalities, staining protocols, scanner types, and tissues significantly affect the appearance of nucleus images, resulting in notable gap between source and target domains \cite{adv_stain,measuring_ds,midog}. If a number of target domain samples are available before testing, one can adopt domain adaptation algorithms to transfer the knowledge learned from the source domain to the target domain \cite{CFAPL,PDAM,MLC}. Unfortunately, in real-world applications, it is usually expensive and time-consuming to collect new training sets for the ever changing target domains; moreover, extra computational cost is required, which is usually unrealistic for the end users. Therefore, it is highly desirable to train a robust nucleus segmentation model that is generalizable to different domains.
	
	In recent years, the research on domain generalization (DG) has attracted wide attention. Most existing DG works are proposed for classification tasks \cite{dg_survey,dg_survey_2} and they can be roughly grouped into data augmentation-, representation learning-, and optimization-based methods. The first category of methods \cite{dg_fsdr,dg_da,dg_ramd,dg_feddg} focus on the way to diversify training data styles and expect the enriched styles cover those appeared in target domains. The second category of methods aim to obtain domain-invariant features. This is usually achieved via improving model architectures \cite{dg_snr,bin,dsu} or introducing novel regularization terms \cite{dg_adv,dg_nc}. The third category of methods \cite{dg_mbdg,dg_l2l,dg_l2d,dg_gm} develop new model optimization strategies, e.g., meta-learning, that improve model robustness via artificially introducing domain shifts during training.
	
	It is a consensus that a generalizable nucleus segmentation model should be robust to image appearance variation caused by the change in staining protocols, scanner types, and tissues, as illustrated in Fig.~\ref{fig:examples}. In this paper, we argue that it is also desirable to be robust to the ratio between foreground (nucleus) and background pixel numbers. This ratio changes the statistics of each feature map channel, and affects the robustness of normalization layers, e.g., instance normalization (IN). We will empirically justify its impact in Section 2.3.
	
	\section{Method}
	\label{sec:method}
	\subsection{Overview}
	In this paper, we adopt a U-Net-based model similar to that in \cite{dcan} as the baseline. It performs both semantic segmentation and contour detection for nucleus instances. The area of each nucleus instance is obtained via subtraction between the segmentation and contour prediction maps \cite{dcan}. Details of the baseline model is provided in the supplementary material. To handle domain variations, we adopt IN rather than batch normalization (BN) in the U-Net model. 
	
	\begin{figure}[tb]
		\centering
		\includegraphics[width=\textwidth]{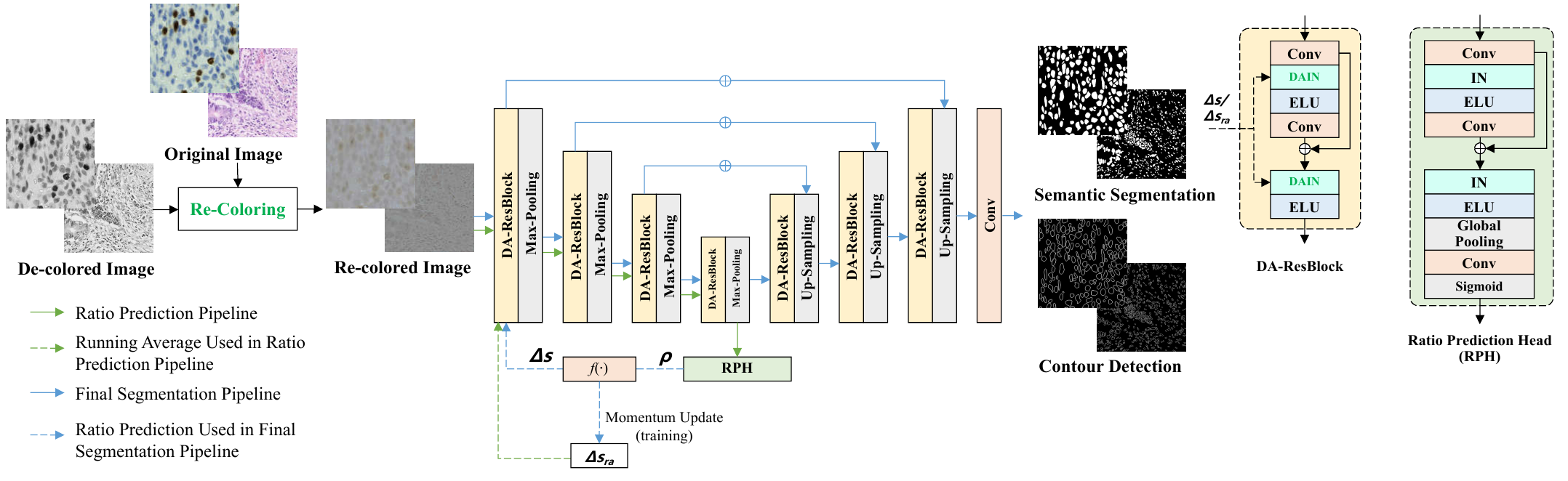}
		\caption{Overview of the DARC model. The whole model is trained in an end-to-end manner. DARC first re-colors each image to relieve the impact caused by different image acquisition conditions. The re-colored image is then fed into the U-Net encoder and the ratio prediction head. This head predicts the foreground-background ratio $\boldsymbol{\rho}$. Then, the re-colored image is fed into DARC again with $\boldsymbol{\rho}$ for final prediction. For simplicity, we only illustrate the data-flow of the first DA-ResBlock in details.}
		\label{fig:overview}
	\end{figure}
	
	Our proposed Distribution-Aware Re-Coloring model (DARC) is illustrated in Fig. 2. Compared with the baseline, DARC replaces the IN layers with the proposed Distribution-Aware Instance Normalization (DAIN) layers. DARC first re-colors each image to relieve the influence caused by image acquisition conditions. The re-colored image is then fed into the U-Net encoder and the ratio prediction head. This head predicts the ratio between foreground and background pixel numbers. With the predicted ratio, the DAIN layers can estimate feature statistics more robustly and facilitate more accurate nucleus segmentation.
	
	\subsection{Nucleus Image Re-Coloring}
	We propose the Re-Coloring (RC) method to overcome the color change in different domains. Specifically, given a RGB image $\boldsymbol{I}$, e.g., an H$\&$E or IHC stained image, we first obtain its grayscale image $\boldsymbol{I}_g$. We then feed $\boldsymbol{I}_g$ into a simple module $T$ that consists of a single residual block and a $1\times1$ convolutional layer with output channel number of 3. In this way, we obtain an initial re-colored image $\boldsymbol{I}_r$. 
	
	However, de-colorization results in the loss of fine-grained textures and may harm the segmentation accuracy. To handle this problem, we compensate $\boldsymbol{I}_r$ with the original semantic information contained in $\boldsymbol{I}$. Recent works \cite{resort} show that semantic information can be reflected via the order of pixels according to their gray value. Therefore, we adopt the Sort-Matching algorithm \cite{sortmatch} to combine the semantic information in $\boldsymbol{I}$ with the color values in $\boldsymbol{I}_r$. Details of RC is presented in Alg.~\ref{alg:rc}, in which $Sort$ and $ArgSort$ denote channel-wisely sorting the values and obtaining the sorted values and indices respectively, and $AssignValue$ denotes re-assembling the sorted values  according to the provided indices. Details of the module $T$ are included in the supplementary material.
	
	\begin{algorithm}[tb]
		\caption{Re-Coloring}
		\label{alg:rc}
		\begin{algorithmic}[1]
			\REQUIRE ~~\\
			The input RGB image $\boldsymbol{I} \in R^{H \times W\times 3}$;\\
			The module $T$ whose input and output channel numbers are 1 and 3, respectively;
			\ENSURE ~~\\
			The re-colored image $\boldsymbol{I}_o \in R^{H \times W\times 3}$;
			\STATE De-colorizing $\boldsymbol{I}$ to obtain ${\boldsymbol{I}_g}$;
			\STATE $\boldsymbol{I}_r \gets T(\boldsymbol{I}_g)$
			\STATE Reshaping $\boldsymbol{I}$ and ${\boldsymbol{I}_r}$ to $R^{HW\times 3}$
			\STATE $\boldsymbol{SortIndex} \gets ArgSort(ArgSort(\boldsymbol{I}))$
			\STATE $\boldsymbol{SortValue} \gets Sort(\boldsymbol{I}_r)$
			\STATE $\boldsymbol{I}_o \gets AssignValue(\boldsymbol{SortIndex}, \boldsymbol{SortValue})$
			\RETURN $\boldsymbol{I}_o$
		\end{algorithmic}
	\end{algorithm}

	\begin{table}[tb]
		\centering
		\caption{Evaluation on the impact of foreground-background ratio to model performance. Both training and testing samples are obtained from CPM17. $B$ denotes the background expansion factor, which directly affects the foreground-background ratio.}
		\label{tab:cd}
		\begin{tabular}{p{40pt}<{\centering} | p{40pt}<{\centering} | p{40pt}<{\centering} | p{40pt}<{\centering} | p{40pt}<{\centering}}
			\hline
			$B$ & 1 & 2 & 4 & 6 \\
			\hline
			AJI  & 65.11 & 59.11 & 53.41 & 54.13 \\
			Dice & 86.14 & 84.05 & 80.97 & 79.75 \\
			\hline
		\end{tabular}
	\end{table}
	
	Via RC, the original fine-grained structure information from $\boldsymbol{I}_g$ is recovered in $\boldsymbol{I}_r$. In this way, the re-colored image is advantageous in two aspects. First, the appearance difference between pathological images caused by the change in scanners and staining protocols is eliminated. Second, the re-colored image preserves fine-grained structure information, enabling precise instance segmentation to be possible. 
	
	\subsection{Distribution-Aware Instance Normalization}
	
	Due to dramatic domain gaps, feature statistics may differ significantly between domains \cite{midog,adv_stain,measuring_ds}, which means that feature statistics obtained from the source domain may not apply to the target domain. Therefore, existing DG works usually replace BN with IN for feature normalization \cite{bin,dg_survey}. However, for dense-prediction tasks like semantic segmentation or contour detection, adopting IN alone cannot fully address the feature statistics variation problem. This is because feature statistics are also relevant to the ratio between foreground and background pixel numbers. Specifically, an image with more nucleus instances produces more responses in feature maps and thus higher feature statistic values, and vice versa. The difference in this ratio causes interference to nucleus segmentation.
	
	To verify the above viewpoint, we evaluate the baseline model under different foreground-background ratios. Specifically, we first remove the foreground pixels via in-painting \cite{inpaint}, and then pad the original testing images with the obtained background patches. We adopt $B$ to denote the ratio between the size of the obtained new image and the original image size. Compared with the original images, the new images have the same foreground regions but more background pixels, and thus have different foreground-background ratios. Finally, we evaluate the performance of the baseline model with different $B$ values. Experimental results are presented in Table~\ref{tab:cd}. It is shown that the value of $B$ affects the model performance significantly.
	
	The above problem is common in nucleus segmentation because pathological images from different organs or tissues tend to have significantly different foreground-background ratios. However, this phenomenon is often ignored in existing research. To handle this problem, we propose the Distribution-Aware Instance Normalization (DAIN) method to re-estimate feature statistics that account for different ratios of foreground and background pixels. Details of DAIN is presented in Alg.~\ref{alg:dain}. The structures of $E_{\boldsymbol{\mu}}$ and $E_{\boldsymbol{\delta}}$ are included in the supplemental materials.
	
	\begin{algorithm}[t]
		\caption{Distribution-Aware Instance Normalization}
		\label{alg:dain}
		\begin{algorithmic}[1]
			\REQUIRE ~~\\
			Original feature maps $\boldsymbol{X} \in R^{H \times W\times C}$. The $C$-dimensional feature vector on its pixel $(i,j)$ is denoted as $\boldsymbol{x}_{ij}$;\\
			The modules $E_{\boldsymbol{\mu}}$ and $E_{\boldsymbol{\delta}}$ that re-estimate feature statistics;\\
			$\boldsymbol{\Delta s_{ra}}\in R^{1 \times 1\times C}$ that is obtained via running mean of $\boldsymbol{\Delta s}$ in the training stage;\\
			The momentum factor $\alpha$ used to update $\boldsymbol{\Delta s_{ra}}$;\\
			(Optional) $\boldsymbol{\Delta s}=f(\boldsymbol{\rho})$;
			\ENSURE ~~\\
			Normalized feature maps $\boldsymbol{Y} \in R^{H \times W\times C}$;
			\STATE $\boldsymbol{\mu} \gets \frac{1}{HW}\sum\limits_{i=1}^{H}\sum\limits_{j=1}^{W}{\boldsymbol{x}_{ij}}$
			\STATE $\boldsymbol{\delta^2} \gets \frac{1}{HW}\sum\limits_{i=1}^{H}\sum\limits_{j=1}^{W}{(\boldsymbol{x_{ij}}-\boldsymbol{\mu})^2}$
			
			\IF {$\boldsymbol{\Delta s}$ is given}
			\STATE // Using $\boldsymbol{\Delta s}$ to re-estimate feature statistics for final segmentation
			\STATE $\boldsymbol{\mu'}, \boldsymbol{\delta'} \gets E_{\boldsymbol{\mu}}(\boldsymbol{\mu}, \boldsymbol{\delta}, \boldsymbol{\Delta s}), E_{\boldsymbol{\delta}}(\boldsymbol{\mu}, \boldsymbol{\delta}, \boldsymbol{\Delta s})$
			\IF {$Training$}
			\STATE // Updating $\boldsymbol{\Delta s_{ra}}$ during training
			\STATE $\boldsymbol{\Delta s_{ra}} \gets (1-\alpha)\boldsymbol{\Delta s_{ra}}+\alpha\boldsymbol{\Delta s}$
			\ENDIF
			\ELSE
			\STATE // Using $\boldsymbol{\Delta s_{ra}}$ to re-estimate feature statistics for ratio prediction
			\STATE $\boldsymbol{\mu'}, \boldsymbol{\delta'} \gets E_{\boldsymbol{\mu}}(\boldsymbol{\mu}, \boldsymbol{\delta}, \boldsymbol{\Delta s_{ra}}), E_{\boldsymbol{\delta}}(\boldsymbol{\mu}, \boldsymbol{\delta}, \boldsymbol{\Delta s_{ra}})$
			\ENDIF
			\STATE $\boldsymbol{Y} \gets (\boldsymbol{X}-\boldsymbol{\mu'}) / \boldsymbol{\delta'}$
			\RETURN $\boldsymbol{Y}$
		\end{algorithmic}
	\end{algorithm}
	
	As shown in Fig. 2, to obtain the foreground-background ratio $\boldsymbol{\rho}$ of one input image, we first feed it to the model encoder with $\boldsymbol{\Delta s_{ra}}$ as the additional input. $\boldsymbol{\Delta s_{ra}}$ acts as pseudo residuals of feature statistics and is obtained in the training stage via averaging $\boldsymbol{\Delta s}$ in a momentum fashion. The output features by the encoder are used to predict the foreground-background ratio $\boldsymbol{\rho}$ with a Ratio-Prediction Head (RPH). $\boldsymbol{\rho}$ is then utilized to estimate the residuals of feature statistics: $\boldsymbol{\Delta s}=f(\boldsymbol{\rho})$. Here, $f$ is a $1\times1$ convolutional layer that transforms $\boldsymbol{\rho}$ to a feature vector whose dimension is the same as the target layer's channel number. After that, the input image is fed into the model again with  $\boldsymbol{\Delta s}$ as additional input and finally makes more accurate predictions. 
	
	The training of RPH requires an extra loss term $L_{rph}$, which is formulated as bellow:
	
	\begin{equation}\label{eq:dloss}
		L_{rph}=L_{BCE}(\boldsymbol{\rho}, \boldsymbol{\rho_g}) + L_{MSE}(f(\boldsymbol{\rho}), f(\boldsymbol{\rho_g})),
	\end{equation}
	where $\boldsymbol{\rho_g}$ denotes the ground truth foreground-background ratio, and $L_{BCE}$ and $L_{MSE}$ denote the binary cross entropy loss and the mean squared error, respectively.
	
	\section{Experiments}
	\label{sec:exps}
	\subsection{Datasets}
	The proposed method is evaluated on four datasets, including two H$\&$E stained image datasets CoNSeP \cite{hovernet} and CPM17 \cite{cpm17} and two IHC stained datasets DeepLIIF \cite{deepliif} and BC-DeepLIIF \cite{deepliif,bcd}. \textbf{CoNSeP} \cite{hovernet} contains 28 training and 14 validation images, whose sizes are $1000\times1000$ pixels. The images are extracted from 16 colorectal adenocarcinoma WSIs, each of which belongs to an individual patient, and scanned with an Omnyx VL120 scanner within the department of pathology at University Hospitals Coventry and Warwickshire, UK. \textbf{CPM17} \cite{cpm17} contains 32 training and 32 validation images, whose sizes are $500\times500$ pixels. The images are selected from a set of Glioblastoma Multiforme, Lower Grade Glioma, Head and Neck Squamous Cell Carcinoma, and non-small cell lung cancer whole slide tissue images. \textbf{DeepLIIF} \cite{deepliif} contains 575 training and 91 validation images, whose sizes are $512\times512$ pixels. The images are extracted from the slides of lung and bladder tissues. \textbf{BC-DeepLIIF} \cite{deepliif,bcd} contains 385 training and 66 validation Ki67 stained images of breast carcinoma, whose sizes are $512\times512$ pixels.
	
	\subsection{Implementation Details}
	In the training stage, patches of size $224\times224$ pixels are randomly cropped from the original samples. During training, the batch size is 4 and the total number of training iterations is 40,000. We use Adam algorithm for optimization, and the learning rate is initialized as $1e^{-3}$, which is gradually decreased to $1e^{-5}$ during training. We adopt the standard augmentation, like image color jittering and Gaussian blurring. In all experiments, the segmentation and contour detection predictions are penalized using the binary cross entropy loss.
	
	\begin{table}[tb]
		\centering
		\caption{Comparisons in generalization performance on nucleus segmentation datasets. Results in each column are related to models trained on one domain and evaluated on the other three unseen domains. Methods marked by * are proposed in this paper. Results are in percentages. }\label{tab:cmp_all}
		\resizebox{1.0\textwidth}{!}{
			\begin{tabular}{p{75pt}<{\centering} | p{27pt}<{\centering} p{27pt}<{\centering} | p{27pt}<{\centering} p{27pt}<{\centering} | p{27pt}<{\centering} p{27pt}<{\centering} | p{27pt}<{\centering} p{27pt}<{\centering} | p{27pt}<{\centering} p{27pt}<{\centering}}
				\multirow{2}*{Methods} & \multicolumn{2}{c|}{CoNSeP} & \multicolumn{2}{c|}{CPM17} & \multicolumn{2}{c|}{DeepLIIF} & \multicolumn{2}{c|}{BC-DeepLIIF} & \multicolumn{2}{c}{Average}\\
				& AJI & Dice & AJI & Dice & AJI & Dice & AJI & Dice & AJI & Dice \\
				\hline
				Baseline (BN)  & 16.67 & 24.10 & 33.30 & 61.18 & 08.42 & 38.17 & 21.27 & 39.92 & 19.92 & 40.84 \\
				Baseline (IN)  & 32.13 & 48.67 & 33.94 & 65.83 & 41.48 & 67.17 & 21.52 & 37.49 & 32.27 & 54.79 \\
				BIN \cite{bin} & 21.54 & 34.33 & 37.06 & 67.63 & 23.51 & 49.49 & 26.15 & 44.42 & 27.01 & 48.97 \\
				DSU \cite{dsu} & 21.42 & 34.66 & 39.12 & 66.55 & 27.21 & 55.10 & 25.09 & 41.83 & 28.21 & 49.53 \\
				SAN \cite{san} & 27.91 & 46.72 & 33.69 & 65.66 & 27.57 & 53.09 & 22.17 & 38.38 & 27.84 & 50.96 \\
				AmpNorm \cite{fn_a,fn_b} & 35.52 & 55.89 & 33.39 & 58.69 & 39.91 & 66.58 & 23.79 & 37.81 & 33.15 &  54.74 \\
				StainNorm \cite{stainnorm} & 41.06 & 60.81 & 32.75 & 64.68 & 38.55 & 63.95 & 25.41 & 43.81 & 34.44 &  58.11 \\
				StainMix \cite{stainmix} & 34.22 & 51.07 & 35.05 & 65.49 & 38.48 & 64.92 & 26.88 & 45.62 & 33.66 &  56.78 \\
				TENT (BN) \cite{dg_tent} & 38.61 & 58.11 & 35.04 & 64.62 & 33.77 & 59.76 & 23.55 & 40.91 & 32.74 & 55.85 \\
				TENT (IN) \cite{dg_tent} & 32.34 & 48.87 & 33.24 & 65.73 & 42.08 & 66.87 & 22.38 & 38.04 & 32.51 & 54.88 \\
				EFDMix \cite{resort}& 40.13 & 58.74 & 33.29 & 65.25 & 39.06 & 64.60 & 25.92 & 42.38 & 34.60 & 57.74 \\
				RC (IN)*            & 37.21 & 57.53 & 36.98 & 67.71 & 35.53 & 62.03 & 24.98 & 42.25 & 33.68 & 57.38 \\
				DAIN*          & 33.86 & 50.08 & 30.62 & 64.64 & 37.93 & 65.56 & 31.20 & 53.15 & 33.40 & 58.87 \\
				DAIN w/o Ratio*& 27.37 & 40.35 & 33.25 & 65.05 & 40.21 & 66.82 & 29.16 & 48.30 & 32.50 & 55.13 \\
				DARC$_{all}$*       & 38.18 & 57.27 & 34.44 & 66.11 & 39.10 & 67.07 & 31.64 & 53.81 & 35.84 & 61.06 \\
				DARC$_{enc}$*       & 40.04 & 58.73 & 35.60 & 66.50 & 40.11 & 68.23 & 32.56 & 53.86 & \textbf{37.08} & \textbf{61.83} \\
			\end{tabular}
		}
	\end{table}
	
	\begin{table}[tb]
		\centering
		\scriptsize
		\caption{Complexity comparison between the baseline model and DARC. }\label{tab:ca}
		\begin{tabular}{p{60pt}<{\centering} | p{60pt}<{\centering} p{60pt}<{\centering}}
			Models & \makecell{$\#$Parameters \\(M)} & \makecell{Inference Time \\(s/image)} \\
			\hline
			Baseline (IN) & 5.03 & 0.0164 \\
			DARC$_{enc}$ & 5.47 & 0.0253\\
			\hline
		\end{tabular}
	\end{table}
	
	\subsection{Experimental Results and Analyses}
	In this paper, the models are compared using the AJI \cite{kumar} and Dice scores. In the experiments, models trained on one of the datasets will be evaluated on the three unseen ones. To avoid the influence of the different sample numbers of the datasets, we calculate the average scores within each unseen domain respectively and then average them across domains.
	
\begin{figure}[tb]
		\centering
		\includegraphics[width=\textwidth]{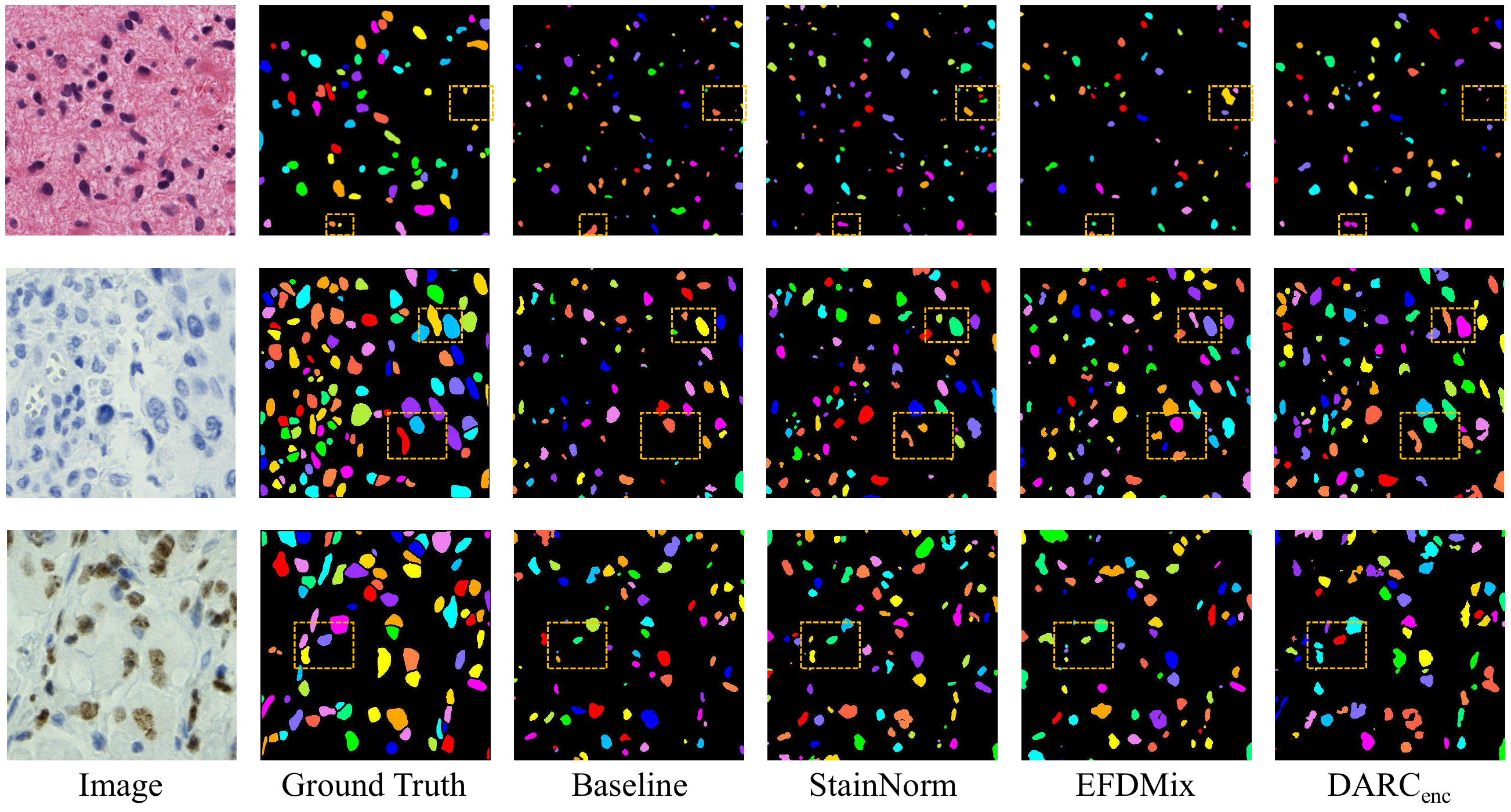}
		\caption{Qualitative comparisons between Different Models.}\label{fig:res_examples}
	\end{figure}

	In this paper, we re-implement some existing popular domain generalization algorithms for comparisons under the same training conditions. Specifically, we re-implement the TENT \cite{dg_tent}, BIN \cite{bin}, DSU \cite{dsu}, Frequency Amplitude Normalization (AmpNorm) \cite{fn_a,fn_b}, SAN \cite{san} and EFDMix \cite{resort}. We also evaluate the stain normalization \cite{stainnorm} and stain mix-up \cite{stainmix} methods that are popular in pathological image analysis. Their performances are presented in Table~\ref{tab:cmp_all}. DARC$_{all}$ replaces all normalization layers with DAIN, while DARC$_{enc}$ replaces the normalization layers in the encoder with DAIN and uses BN in its decoder. As shown in Table~\ref{tab:cmp_all}, DARC$_{enc}$ achieves the best average performance among all methods. Specifically, DARC$_{enc}$ improves the baseline model's average AJI and Dice scores by $4.81\%$ and $7.04\%$. Compared with the other domain generalization methods, DAIN, DAIN w/o Ratio, DARC$_{all}$ and DARC$_{enc}$ achieve impressive performances on BC-DeepLIIF, which justify that re-estimating the instance-wise statistics is important for improving the domain generalization ability of models trained on BC-DeepLIIF. Qualitative comparisons are presented in Fig.~\ref{fig:res_examples}. Moreover, the complexity analysis between the baseline model and DARC$_{enc}$ is presented in Table~\ref{tab:ca}.
	
	We separately evaluate the effectiveness of RC and DAIN, and present the results in Table~\ref{tab:cmp_all}. Also, we train a variant model without foreground-background ratio prediction, which is denoted as `DAIN w/o Ratio' in Table~\ref{tab:cmp_all}. Compared with the baseline model, RC improves the average AJI and Dice scores by 1.41$\%$ and 2.59$\%$, and DAIN improves the average AJI and Dice scores by 1.13$\%$ and 4.08$\%$. Compared with the variant model without foreground-background ratio prediction, DAIN improves the average AJI and Dice scores by 0.90$\%$ and 3.74$\%$. Finally, the combinations of RC and DAIN, i.e., DARC$_{all}$ and DARC$_{enc}$, achieve the best average scores. As shown in Table~\ref{tab:cmp_all}, DARC$_{enc}$ improves DARC$_{all}$ by 1.24$\%$ and 0.77$\%$ on AJI and Dice scores respectively. This is because after the operations by RC and DAIN in the encoder, the obtained feature maps are much more robust to the domain gaps, which enables the decoder to adopt the fixed statistics maintained during training. Moreover, using the fixed statistics is helpful to prevent the decoder from the influence of varied foreground-background ratios on feature statistics.
	
	\section{Conclusion}
	In this paper, we propose the DARC model for generalizable nucleus segmentation. To handle the domain gaps caused by varied image acquisition conditions, DARC first re-colors the input image while preserving its fine-grained structures as much as possible. Moreover, we find that the performance of instance normalization is sensitive to the varied ratios in foreground and background pixel numbers. This problem is well addressed by our proposed DAIN. Compared with existing works, DARC achieves significantly better performance on average across four benchmarks.


\end{document}